\documentclass[twoside,11pt]{article}
\pdfoutput=1

\usepackage{geometry}
\usepackage{natbib}
\usepackage{graphicx}
\usepackage{rotating}
\usepackage{amsmath}
\usepackage{amssymb}

\usepackage{theorem}
\newcommand{\BlackBox}{\rule{1.5ex}{1.5ex}}  

\newcommand{\cbr}[1]{\left\{#1\right\}}

\newcommand{\intset}[1]{\cbr{1..n}}

\usepackage[colorlinks,pdfpagelabels,plainpages=false]{hyperref}
\usepackage{algorithm}
\usepackage{algorithmic}
\usepackage{rotating}

\usepackage{xcolor}
\definecolor{dark-red}{rgb}{0.4,0.15,0.15}
\definecolor{dark-blue}{rgb}{0.15,0.15,0.4}
\definecolor{medium-blue}{rgb}{0,0,0.5}
\hypersetup{
   colorlinks, linkcolor={dark-blue},
   citecolor={dark-blue}, urlcolor={medium-blue}
}

\usepackage{booktabs}
\usepackage{bm}
\usepackage{amsmath}
\usepackage{amssymb}
\usepackage{amsfonts}
\usepackage{mathtools}
\usepackage{comment}
\usepackage{subfigure}

\newcommand{\mbf}[1]{{\boldsymbol{\mathbf{#1}}}}
\renewcommand{\bm}{\mbf}

\usepackage{color}

\usepackage{array}
\newcolumntype{L}[1]{>{\raggedright\let\newline\\\arraybackslash\hspace{0pt}}m{#1}}
\newcolumntype{C}[1]{>{\centering\let\newline\\\arraybackslash\hspace{0pt}}m{#1}}
\newcolumntype{R}[1]{>{\raggedleft\let\newline\\\arraybackslash\hspace{0pt}}m{#1}}

\usepackage{parskip}

\title{Scalable Gaussian Processes for Characterizing \\Multidimensional Change Surfaces}

\author{
  William Herlands \\
  Carnegie Mellon University \\
  herlands@cmu.edu
  \and
  Andrew Wilson \\
  Carnegie Mellon University \\
  andrewgw@cs.cmu.edu
  \and
  Hannes Nickisch \\
  Philips Research Hamburg \\
  hannes@nickisch.org
  \and
  Seth Flaxman \\
  Oxford University \\
  flaxman@stats.ox.ac.uk
  \and
  Daniel Neill \\
  Carnegie Mellon University \\
  neill@cs.cmu.edu
  \and
  Wilbert van Panhuis \\
  University of Pittsburgh \\
  WAV10@pitt.edu
  \and
  Eric Xing \\
  Carnegie Mellon University \\
  epxing@cs.cmu.edu
}

\begin{document}
\date{}  

\maketitle

\begin{abstract} 

\begin{sloppypar}
We present a scalable Gaussian process model for identifying and characterizing smooth multidimensional changepoints, and automatically learning changes in expressive covariance structure. We use Random Kitchen Sink features to flexibly define a \emph{change surface} in combination with expressive spectral mixture kernels to capture the complex statistical structure.  Finally, through the use of novel methods for additive non-separable kernels, we can scale the model to large datasets. We demonstrate the model on numerical and real world data, including a large spatio-temporal disease dataset where we identify previously unknown heterogeneous changes in space and time. 
\end{sloppypar}

\end{abstract}

\section{Introduction}

In human systems we are often confronted with changes or perturbations which may not immediately disrupt an entire system. Instead, changes such as policy interventions take time to affect deeply held habits or trickle through a complex bureaucracy. The dynamics of these changes are non-trivial, with sophisticated spatial distributions, rates, and intensity functions. Using expressive models to fully characterize such changes is essential for making accurate predictions and yielding scientifically relevant results.

Typically, changepoint methods \citep{chernoff1964estimating} model system perturbations as discrete, or near-discrete, changepoints. These points are either identified sequentially using online algorithms, or retrospectively. Here we consider retrospective analysis \citep{brodsky2013nonparametric,chen2011parametric}.

Gaussian processes have been used for changepoint modeling to provide a nonparametric framework. \citet{saatcci2010gaussian} extend the sequential Bayesian Online Changepoint Detection algorithm \citep{adams2007bayesian}, by using a Gaussian process to model temporal covariance  within a particular regime. Similarly, \citet{garnett2009sequential} provide Gaussian processes for sequential changepoint detection with mutually exclusive regimes. These models focus on discrete changepoints, where regimes defined by distinct Gaussian processes change instantaneously at $t=t_0$. While such models may be appropriate for mechanical systems, they do not permit modeling of the complex changes common to many human systems.  

A small collection of pioneering work has briefly considered the possibility of non-discrete Gaussian process change-points \citep{wilson2014covariance,lloyd2014automatic}. Yet these models rely on sigmoid transformations of linear functions which are restricted to fixed rates of change, and are demonstrated exclusively on small, one-dimensional time series data. They cannot expressively characterize non-linear changes or feasibly operate on large multidimensional data.

Applying changepoints to multiple dimensions, such as spatio-temporal data, is theoretically and practically non-trivial, and has thus been seldom attempted. Notable exceptions include \citet{majumdar2005spatio} who consider discrete spatio-temporal changepoints with three additive Gaussian processes: one for $t \leq t_0$, one for $t > t_0$, and one $\forall t$. Alternatively, \citet{nicholls2010building} use a Bayesian onset-field process on a lattice to model the spatio-temporal distribution of human settlement on the Fiji islands.

The limitations of these models reflect a common criticism that Gaussian processes are unable to convincingly respond to changes in covariance structure. We propose addressing this deficiency with an expressive, flexible, and scalable change surface model.

Throughout the paper we refer to \emph{change surfaces} as the multidimensional generalization of changepoints. Unlike the discrete notion of changepoints, a change surface can have a variable rate of change and non-monotonicity in the transition between functional regimes. Additionally, changes can occur heterogeneously across the input dimensions. We formalize the notion of a change surface through our model specification in Section \ref{sec:changepoint_model}.

\subsection{Main contributions}
We introduce a scalable Gaussian process model, which is capable of automatically learning expressive covariance functions, including a sophisticated continuous change surface. We derive scalable inference procedures leveraging Kronecker structure, and a lower bound on the marginal likelihood using the Weyl inequality, as a principled means for scalable kernel learning. Our contributions include:
\begin{enumerate}
\item A non-discrete Gaussian process change surface model over multiple input dimensions. Our model specification learns the change surface from data, enabling it to approximate discrete changes or gradual shifts between regimes. The input can have arbitrary dimension, though we primarily focus our attention on spatio-temporal modeling over 2D space and 1D time.
\item The first scalable Gaussian process changepoint model by using novel Kronecker methods. Modern datasets require methods which can scale to hundreds of thousands of instances.
\item A novel method for estimating the log determinant of additive positive semidefinite matrices using the Weyl inequality. This enables scalable additive Gaussian process models with non-separable kernels in space and time.
\item Random Kitchen Sink features to sample from a Gaussian process change surface. This flexibility permits arbitrary changes which can adapt to heterogeneous effects over multiple dimensions. It also allows us to analytically optimize the entire model.
\item We use logistic functions to normalize the weights on all latent functions (one per regime), thereby providing a very interpretable model. Additionally, we permit arbitrary specification of the change surface parameterization, allowing experts to specify interpretable models for how the change surface behaves over the input space.
\item A novel initialization method for spectral mixture kernels by fitting a Gaussian mixture model to the Fourier transform of the data. This provides good starting values for hyperparameters of expressive stationary kernels, allowing for proper optimization over a multimodal parameter space.
\item A nonparametric Bayesian framework for discovering and characterizing continuous changes in large observational data. We demonstrate our approach on numerical and real world data, including a recently developed public health dataset. We demonstrate how the effect of the measles vaccine introduced in the US in 1963 was spatio-temporally varying. Our model discovers the time frame in which the measles vaccine was introduced, and accurately represents the change in dynamics before and after the introduction, thus providing new insights into the spatial and temporal dynamics of reported disease incidence.
\end{enumerate}

\subsection{Outline}
In the remainder of the paper, section~\ref{sec:GPs} provides background on Gaussian processes. Section~\ref{sec:changepoint_model} describes our change surface model including the weighting, warping, and kernel functions. Section~\ref{sec:Inference} introduces our novel algorithm for approximating the log determinant of additive kernels. Section~\ref{sec:initialization} details our initialization procedure including our new approach for spectral mixture hyperparameter initialization. Section~\ref{sec:experiments} describes our numerical and real-world experiments. Finally, we conclude with summary remarks in section~\ref{sec:conclusions}.

\section{Gaussian Process}
\label{sec:GPs}
Given data $(\bm{y}, \bm{x})$, where $\bm{y}= \{y_1...y_n\}$, are outputs or response variables, and $\bm{x} = \{x_1...x_n\}, x_i \in R^D$ are inputs or covariates, we assume that the responses are generated from the inputs by a latent function with a Gaussian process prior and Gaussian noise, such that $\bm{y} = f(\bm{x}) + \epsilon$, $f(x) \sim GP(m, k)$, $\epsilon \sim \mathcal{N}(0, \sigma_\epsilon)$. A Gaussian process is a nonparametric prior over functions completely specified by mean and covariance functions:
\begin{align}
f(x) &\sim \mathcal{GP}(m(x), k(x,x'))\\
m(x) &= \mathbb{E}[f(x)]\\
k(x,x') &= \text{cov}(f(x), f(x'))
\end{align}
Any finite collection of function values is normally distributed $[f(x_1) ... f(x_p)] \sim \mathcal{N}(\bm{\mu}, K)$ where $\mu_i = m(x_i)$ and $p\times p$ matrix $K_{i,j} = k(x_i,x_j)$. 

In order to learn hyperparameters, we often desire to optimize the marginal likelihood of the data, conditioned on kernel hyperparameters $\theta$, and inputs, $\bm{x}$.
\begin{eqnarray}
p(\bm{y}|\theta, \bm{x}) = \int p(y|f,\bm{x})p(f|\theta) df
\end{eqnarray}
In the case of a Gaussian observation model we can express the log marginal likelihood as,
\begin{equation}
\begin{aligned}
\label{eq:GPloglik}
\log p(\bm{y}|\theta)  \propto - \log|K + \sigma_\epsilon I| - \bm{y}^{\top}(K + \sigma_\epsilon I)^{-1} \bm{y}
\end{aligned}
\end{equation}
We assume familiarity with the basics of Gaussian processes as described by \citet{rasmussen2006gaussian}.

\section{Smooth Change Surface Model}
\label{sec:changepoint_model}

Change surface data consists of latent functions $f_1,\dots,f_r$ defining $r$ regimes in the data. The transition between any two functions is considered a change surface. Were these $r$ functions not mutually exclusive, we could consider an input dependent mixture model such as \citep{wilson2011gaussian},
\begin{eqnarray}
\label{eq:GPRN}
y(x) = w_1(x)f_1(x) + \dots + w_r(x)f_r(x) + \epsilon_n
\end{eqnarray}
where the weighting functions, $w_i(x): R^D \rightarrow R^1$, describe the mixing proportions over the input domain. However, for data with changing regimes we are particularly interested in latent functions that exhibit some amount of mutual exclusivity. 

We induce this partial discretization with a warping function, $\sigma(z): R^1 \rightarrow [0,1]$, which has support over the entire real line but a range which is concentrated towards 0 and 1. Additionally, we choose $\sigma(z)$ such that it produces a convex combination over the weighting functions, $\sum_{i=1}^r \sigma(w_i(x)) = 1$. In this way, each $w_i(x)$ defines the strength of latent $f_i$ over the domain, while $\sigma(z)$ normalizes these weights to induce weak mutual exclusivity.

A natural choice for flexible, smooth change surfaces is the softmax function since it can approximate a Heaviside step function or gradual changes. For $r$ latent functions, the resulting warping function is
\begin{eqnarray}
\label{eq:softmax}
\sigma(w_i(x)) = \text{softmax}(\bm{w}(x))_i = \frac{\exp(w_i(x))}{\sum_{j=1}^r \exp(w_j(x))} \,.
\end{eqnarray}

Our model is thus,
\begin{eqnarray}
\label{eq:y_GP_changepoint}
y(x) = \sigma(w_1(x))f_1(x) + \dots + \sigma(w_r(x))f_r(x) + \epsilon_n
\end{eqnarray}
If we assume Gaussian process priors on all latent functions $f_1(x),\dots,f_r(x)$ we can define $y(x) = f(x) + \epsilon$ where $f(x)$ has a Gaussian process prior with covariance function,
\begin{equation}
\begin{aligned}
\label{eq:k_GP_changepoint}
k(x,x') = \sigma(w_1(x))k_1(x,x')\sigma(w_1(x')) + \\  \dots + \sigma(w_r(x))k_r(x,x')\sigma(w_r(x'))
\end{aligned}
\end{equation}
This assumption does not limit the expressiveness of Eq.~\ref{eq:y_GP_changepoint} since each Gaussian process may be defined with different mean and covariance functions. Indeed, where the data exhibits latent functional change we expect that the latent functions will have correspondingly different hyperparameters even if the kernel forms are identical.

$\sigma(w_1(x))\dots\sigma(w_r(x))$ induce nonstationarity since they are dependent on the input $x$. Thus, even if we use stationary kernels for all $k_i$, our model results in a flexible, nonstationary kernel.

Each $\sigma(w_i(x))$ defines how the coverage of $f_i(x)$ varies over the input domain. Where $\sigma(w_i(x)) \approx 1$, $f_i(x)$ dominates and primarily describes the relationship between $\bm{x}$ and $\bm{y}$, and in cases where there is no $i$ such that $\sigma(w_i(x)) \approx 1$, a number of functions are dominant in defining the relationship between $\bm{x}$ and $\bm{y}$. Since $\sigma(z)$ pushes values towards 1 or 0, the regions with multiple dominant functions are transitory and thus considered change regions. Therefore, we can interpret how the change surface develops and where different regimes dominate by evaluating $\sigma(w(x))$ over the input domain.

\subsection{Design choices for $w(x)$}
\label{sec:wx}

The functional form of $w(x)$ determines how changes can occur in the data, and how many can occur. For example, a linear parametric weighting function,
\begin{eqnarray}
\label{eq:w_linear}
w(x) = \beta_0 + \beta_1^{\top} x \,,
\end{eqnarray}
only permits a single linear change surface in the data. Yet even this simple model is more expressive than discrete changepoints since it permits flexibility in the rate of change and extends to change regions in $R^D$.

In order to develop a general framework we do not require any prior knowledge about the functional form of $w(x)$ and instead assume a Gaussian process prior on $w(x)$. While in principle we could sample from the full Gaussian process prior, this would lead to a non-conjugate model which would thus be less computationally attractive and significantly constrain the ``plug and play'' nature of choices for $\sigma(z)$, $w(x)$, and $K$. Instead, we approximate the Gaussian process with Random Kitchen Sink (RKS) features and analytically derive inference procedures using the log marginal likelihood \citep{lazaro2010sparse}.

\citet{rahimi2007random} demonstrate that if we consider the vector of RKS features which maps the $D$ dimensional input $x$ to an $m$ dimensional feature space,
\begin{eqnarray}
\label{eq:RKS_features}
\phi(x)^{\top} = \sqrt{\frac{2}{m}} [\cos(\omega_i^{\top} x + b_i) ]_{i=1}^m
\end{eqnarray}
then we can approximate any stationary kernel by taking the Fourier transform of $k(x,x') = k(x-x')$,
\begin{eqnarray}
p(\omega) = \frac{1}{2\pi} \int \exp(-j \omega \delta) k(\delta) d\delta
\end{eqnarray}
and putting priors over the parameters of the RKS feature mapping,
\begin{eqnarray}
\omega_i \sim p(\omega) \\
b_i \sim \mbox{Uniform}(0, 2\pi)
\end{eqnarray}
For an RBF kernel where $\Lambda = \mbox{diag}(l_1^2,\dots,l_D^2)$ is a diagonal matrix of length-scales, we sample,
\begin{eqnarray}
\omega_i \sim \mathcal{N}(0, \frac{1}{4\pi^2}\Lambda^{-1})
\end{eqnarray}
Therefore, if we want to place a Gaussian process prior over our weighting functions, $w(x)\sim GP(0,K)$, we can use RKS features to create a compact representation of the kernel \citep{lazaro2010sparse}. For any finite input $\bm{x}$ we know that,
\begin{eqnarray}
g(\bm{x}) \sim \mathcal{N}(0, K)
\end{eqnarray}
Equivalently, we can define parameters $a$ such that,
\begin{eqnarray}
a \sim \mathcal{N}(0, \frac{\sigma_0}{m}I)\\
w(\bm{x}) = \phi(\bm{x})^{\top} a
\end{eqnarray}
which we can write in the explicit RKS feature space representation,
\begin{eqnarray}
w(x_i) = \sum_{i=1}^v a_i \cos(\omega_i^{\top} x + b_i)
\end{eqnarray}
allowing us to sample from $w(x)$ with a finite sum of RKS features. Initialization of hyperparameters $\sigma_0$ and $\Lambda$ is discussed in Section \ref{sec:initialization}.

Experts with domain knowledge can specify a parametric form for $w(x)$ other than RKS features. Such specification can be advantageous, requiring relatively few, highly interpretable parameters to optimize. Additionally, specifying the functional form of $w(x)$ does not require prior knowledge about if, where, or how rapidly changes occur.

\subsection{Design choices for $K$}

Each latent function is specified by a kernel with unique hyperparameters. By design, each $k_i$ may be of a different form. For example, one function may have a Mat\'ern kernel, another a periodic kernel, and a third an exponential kernel. Such specification is useful when domain knowledge provides insight into the covariance structure of the various regimes. 

In order to maintain maximal generality and expressivity, we develop the model using spectral mixture kernels \citep{wilson2013gaussian}
where $k_{SM}(\tilde{x},\tilde{x}') = $
\begin{equation*}
\label{SM_kernel}
\sum_{q=1}^Q \omega_q cos(2\pi (\tilde{x}-\tilde{x}')^{\top}m_q) \prod_{p=1}^P\exp(-2\pi^2 (\tilde{x}_p - \tilde{x}'_p)^2 v_q^{(p)}) \,,
\end{equation*}
where $\tilde{x}\in R^P$ and $\Sigma_q = \mbox{diag}(v_q^{(1)},\dots,v_q^{(P)})$ is a diagonal covariance matrix for multidimensional inputs. With a sufficiently large $Q$, spectral mixture kernels can approximate any stationary kernel, providing the flexibility to capture complex patterns over multiple dimensions. These kernels have been used in pattern prediction, outperforming complex combinations of standard stationary kernels \citep{wilson2014fast}.

Using spectral mixture kernels extends previous work on Gaussian processes changepoint modeling which has been restricted in practice to RBF \citep{saatcci2010gaussian,garnett2009sequential} or exponential kernels \citep{majumdar2005spatio}. Expressive covariance functions are particularly important with multidimensional and spatio-temporal data where the dynamics are complex and unknown a priori. While most Gaussian process models provide the theoretical flexibility to choose any kernel, the practical mechanics of initializing and fitting more expressive kernels is a challenging problem. We describe an initialization procedure in Section \ref{sec:initialization} which we hope can enable other models to exploit expressive kernels as well.

\section{Scalable inference}
\label{sec:Inference}

Analytic optimization and inference requires computation of the log marginal likelihood (Eq.~\ref{eq:GPloglik}). Yet calculating the inverse and log determinant of $n \times n$ covariance matrices requires $O(n^3)$ computations and $O(n^2)$ memory  \citep{rasmussen2006gaussian}, 
which is impractical for large datasets. Recent advances in scalable Gaussian processes have reduced this computational burden by exploiting Kronecker structure under two assumptions. One, the inputs lie on a grid formed by a Cartesian product, $x \in X = X^{(1)} \times...\times X^{(D)}$. Two, the kernel is multiplicative across each dimension. The assumption of separable, multiplicative kernels is commonly employed in spatio-temporal Gaussian process modeling \citep{martin1990use,majumdar2005spatio,flaxman2015fast}. Under these assumptions, the $n\times n$ covariance matrix $K=K_1\otimes \dots \otimes K_D$, where each $K_d$ is $n_d \times n_d$ such that $\prod_1^D n_d = n$.

Using efficient Kronecker algebra, \citet{saatcci2012scalable} calculates the inverse and log determinant calculations in $O(Dn^{\frac{D+1}{D}})$ operations using $O(Dn^{\frac{2}{D}})$ memory. Furthermore, \citet{wilson2014fast} extends the Kronecker methods for incomplete grids.

Yet for an additive kernel such as that needed for change surface modeling (Eq.~\ref{eq:k_GP_changepoint}),
calculating the inverse and log determinant is no longer feasible using Kronecker algebra as in \citet{saatcci2012scalable} because the sum of the matrix Kronecker products does not decompose as a single Kronecker product. Instead, calculations involving the inverse can be efficiently carried out using linear conjugate gradients as in \citet{flaxman2015fast} because the key subroutine is matrix-vector multiplication and the sum of Kronecker products can be efficiently multiplied by a vector.

However, there is no exact method for efficient computation of the log determinant of the sum of Kronecker products. Instead, \citet{flaxman2015fast} upper bound the log determinant using the Fiedler bound \citep{fiedler1971bounds} which says that for $n \times n$ Hermitian matrices $A$ and $B$ with sorted eigenvalues $\alpha_1,\dots,\alpha_n$ and $\beta_1,\dots,\beta_n$ respectively,
\begin{eqnarray}
\label{eq:fiedler}
\log(| A + B |) \leq \sum_{i=1}^n \log(\alpha_i + \beta_{n-i+1})
\end{eqnarray}
While this yields fast, $O(n)$ computation, the Fiedler bound does not generalize for more than two matrices. 

Instead, we bound the log determinant of the sum of multiple covariance matrices using Weyl's inequality \citep{weyl1912asymptotische} which states that for $n\times n$ Hermitian matrices, $M = A + B$, with sorted eigenvalues $\mu_1,\dots,\mu_n$, $\alpha_1,\dots,\alpha_n$, and $\beta_1,\dots,\beta_n$, respectively,
\begin{eqnarray}
\label{eq:fiedler}
\mu_{i+j-1} \leq \alpha_i + \beta_j
\end{eqnarray}
Since $\log(| A + B |) = \log(|M|) = \sum_{i=1}^n \log(\mu_i)$ we can bound the log determinant by $\sum_{i+j-1=1}^n \log(\alpha_i + \beta_j)$. Furthermore, we can use the Weyl bound iteratively over pairs of matrices to bound the sum of $r$ covariance matrices $K_1,\dots,K_r$.

As the bound indicates, there is flexibility in the choice of which eigenvalue pair $\{\alpha_i, \beta_j\}$ to sum in order to bound $\mu_{i+j-1}$. One might be tempted to minimize over all possible pairs for each of the $n$ eigenvalues of $M$ in order to obtain the tightest bound on the log determinant. Unfortunately, this requires $O(n^2)$ computations. Instead we explore two possible alternatives:
\begin{enumerate}
\item For each $\mu_{i+j-1}$ we choose the ``middle'' pair such that $i = j$ when possible, and $i = j+1$ otherwise. This heuristic requires $O(n)$ computations.
\item We employ a greedy search by using the previous $i'$ and $j'$ to choose the minimum of $2s$ pairs of eigenvalues $\{\alpha_i, \beta_j\}_{i=i'-s}^{i=i'+s}$. When $s=0$ this corresponds to the middle heuristic. When $s=\frac{n}{2}$ this corresponds to the exact Weyl bound. The greedy search requires $O(2sn)$ computations.
\end{enumerate}
In addition to bounding the sum of kernels, we must also deal with the scaling functions, $\sigma(w_i(x))$. We can rewrite Eq.~\ref{eq:k_GP_changepoint} in matrix notation,
\begin{eqnarray}
\label{eq:k_GP_changepoint_matrix}
K = S_1K_1S_1' +  \dots + S_rK_rS_r'
\end{eqnarray}
where $S_i = \mbox{diag}(\sigma(w_i(x)))$ and $S_i' = \mbox{diag}(\sigma(w_i(x')))$. Employing the bound on eigenvalues of matrix products \citep{bhatia2013matrix},
\begin{eqnarray}
\mbox{sort}(\mbox{eig}(A*B)) \leq \mbox{sort}(\mbox{eig}(A))*\mbox{sort}(\mbox{eig}(B))
\end{eqnarray}
we can bound the log determinant of $K$ in Eq.~\ref{eq:k_GP_changepoint_matrix} with a Weyl approximation over $[\{s_{i,l} * k_{i,l} * s_{i,l}'\}_{l=1}^n]_{i=1}^r$ where $s_{i,l}$ is the $l^{th}$ largest eigenvalue of $S_i$ and $k_{i,l}$ is the $l^{th}$ largest eigenvalue of $K_i$

We empirically evaluate the exact Weyl bound, middle heuristic, and greedy search with $s=40$ for our model using synthetic data (generated according to the procedure in Section \ref{sec:numerical_exp}). We compare these results against the Fiedler bound (in the case of two kernels), and a recently proposed method for estimating the log determinant using Chebyshev polynomials coupled with stochastic Hutchinson trace approximation \citep{han2015large}.
\begin{figure}[h]
\begin{centering}

 \includegraphics[width=0.7\textwidth]{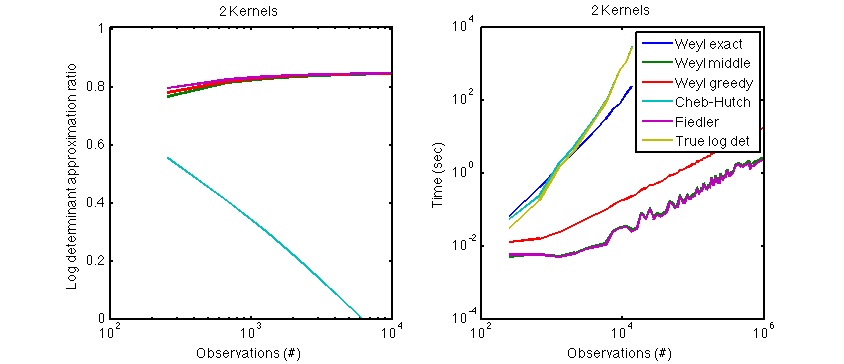}

\caption{Left plot shows the ratio of approximations to the true log determinant of 2 additive kernels. Right plot shows the time to compute each approximation and the true log determinant of 2 additive kernels.}
\label{fig:weyl_2kernels}
\end{centering}
\end{figure}
\begin{figure}[h]
\begin{centering}

 \includegraphics[width=0.7\textwidth]{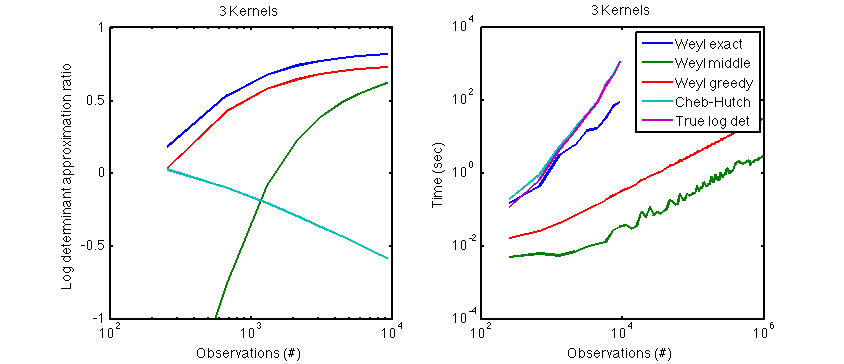}

\caption{Left plot shows the ratio of approximations to the true log determinant of 3 additive kernels. Right plot shows the time to compute each approximation and the true log determinant of 3 additive kernels.}
\label{fig:weyl_3kernels}
\end{centering}
\end{figure}
Figures \ref{fig:weyl_2kernels} and \ref{fig:weyl_3kernels} depict the ratio of each approximation to the true log determinant, and the time to compute each approximation over increasing number of observations for 2 and 3 kernels. We note that all Weyl and Fiedler approximations converge to $\approx 0.8$ of the true log determinant, which was negative in the experiments. While the exact Weyl bound scales poorly, as expected, both approximate Weyl bounds scale well. In practice, we use the middle heuristic since it provides the fastest results.

\section{Initialization}
\label{sec:initialization}

Since our model uses expressive spectral mixture kernels and flexible RKS features, the parameter space is highly multimodal. Therefore, it is essential to initialize the model hyperparameters appropriately.  Below we present a method where we first initialize the $w(x)$ RKS features and then use those values in a novel initialization method for the spectral mixture kernels.

To initialize $w(x)$ we simplify our model and assume that each $k_i$ is an RBF kernel. Using the procedure in Algorithm \ref{alg:init_wx} we test $g$ possible $w(x)$ functions by drawing the hyperparameters $a$, $\omega$, and $b$ from their respective prior distributions (Section \ref{sec:wx}). We set reasonable values of $\Lambda = (\frac{range(x)}{2})^2$, $\sigma_0=\mbox{std}(y)$, and $\sigma_n= \frac{\mbox{mean}(|y|)}{10}$.

For each $w(x)$, we sample $h$ possible sets of hyperparameters for the RBF kernels and select the best set via maximum marginal likelihood. Then we run an abbreviated optimization procedure over each of the $g$ sets of $w(x)$ and RBF hyperparameters and select the best set via marginal likelihood. Finally, we optimize all the resulting parameters until convergence.
\begin{algorithm}
\caption{Initialize RKS $w(x)$ by optimizing a simplified model with RBF kernels}
\label{alg:init_wx}
\begin{algorithmic}[1]
\FOR{$i=1:g$}
\STATE Draw $a$, $\omega$, $b$ for RKS features in $w(x)$
\STATE Draw $h$ random values for RBF kernels. Choose the best with maximum marginal likelihood
\STATE Partial optimization of $w(x)$ and RBF kernels
\ENDFOR
\STATE Choose the best set of hyperparameters with maximum marginal likelihood
\STATE Optimize all hyperparameters until convergence
\end{algorithmic}
\end{algorithm}

In order to initialize the spectral mixture kernels, we use the optimized $w(x)$ from above to define the subset $\{x : \sigma(w_i(x)) > 0.5\}$ where the latent $f_i(x)$ is dominant. For each $f_i$ we then take a Fourier transform of $y(x)$ over each dimension of $\{x : \sigma(w_i(x)) > 0.5\}$ to obtain the empirical spectrum in that dimension. Note that we consider each dimension of $x$ individually since we have a multiplicative Q-component spectral mixture kernel over each dimension. Since spectral mixture kernels model the spectral density with $Q$ Gaussians on $R^1$, we fit a 1D Gaussian mixture model,
\begin{eqnarray}
\label{eq:GMM}
p(x) = \sum_{q=1}^Q \phi_q \mathcal{N}(\mu_q, \sigma_q)
\end{eqnarray}
to the the empirical spectrum for each dimension. Using the learned mixture model we initialize the parameters of our spectral mixture kernels for $f_i(x)$ as detailed in line 8 of Algorithm \ref{alg:init_SM}.

Finally, we use the initialized $w(x)$ and spectral mixture kernel hyperparameters and jointly optimize the model using marginal likelihood and standard gradient techniques \citep{rasmussen2010gaussian}.
\begin{algorithm}
\caption{Initialize spectral mixture kernels}
\label{alg:init_SM}
\begin{algorithmic}[1]
\STATE Use $w(x)$ output from Algorithm \ref{alg:init_wx}
\FOR{$k_i:i=1:r$} 
\FOR{$d=1:D$}
\FOR{Each $x_d|x_{-d}$}
\STATE Sample $s \sim | FFT(sort(y(x_d|x_{-d}))) |^2$ [empirical spectrum for dimension $d$]
\ENDFOR
\STATE Fit Q component 1D GMM to all samples
\STATE Initialize $\omega_q = std(y) * \phi_q$; $m_q = \mu_q$; $v_q = \sigma_q$
\ENDFOR
\ENDFOR
\end{algorithmic}
\end{algorithm}

\section{Experiments}
\label{sec:experiments}

We test our model with both numerical and real world data. There do not exist standard datasets for evaluating spatio-temporal changepoint models. For example, \citet{majumdar2005spatio} used simulations to demonstrate the effectiveness of their model. Therefore, we apply our method on a standard 1D changepoint dataset, synthetic data, and a newly available spatio-temporal disease dataset.

\subsection{Numerical Experiments}
\label{sec:numerical_exp}
We generate a $50\times 50$ grid of synthetic data by drawing independently from two latent functions. Each function is characterized by a 2D RBF kernel with different length-scales and variances. The synthetic change surface between the functions is defined by $\sigma(w_{poly}(x))$ where $w_{poly}(x)=\sum_{i=0}^3 \beta_i^T x^i$, $\beta_i \sim \mathcal{N}(0,3 I_D)$.

\begin{figure}[h]
\begin{centering}
 \includegraphics[width=0.6\textwidth]{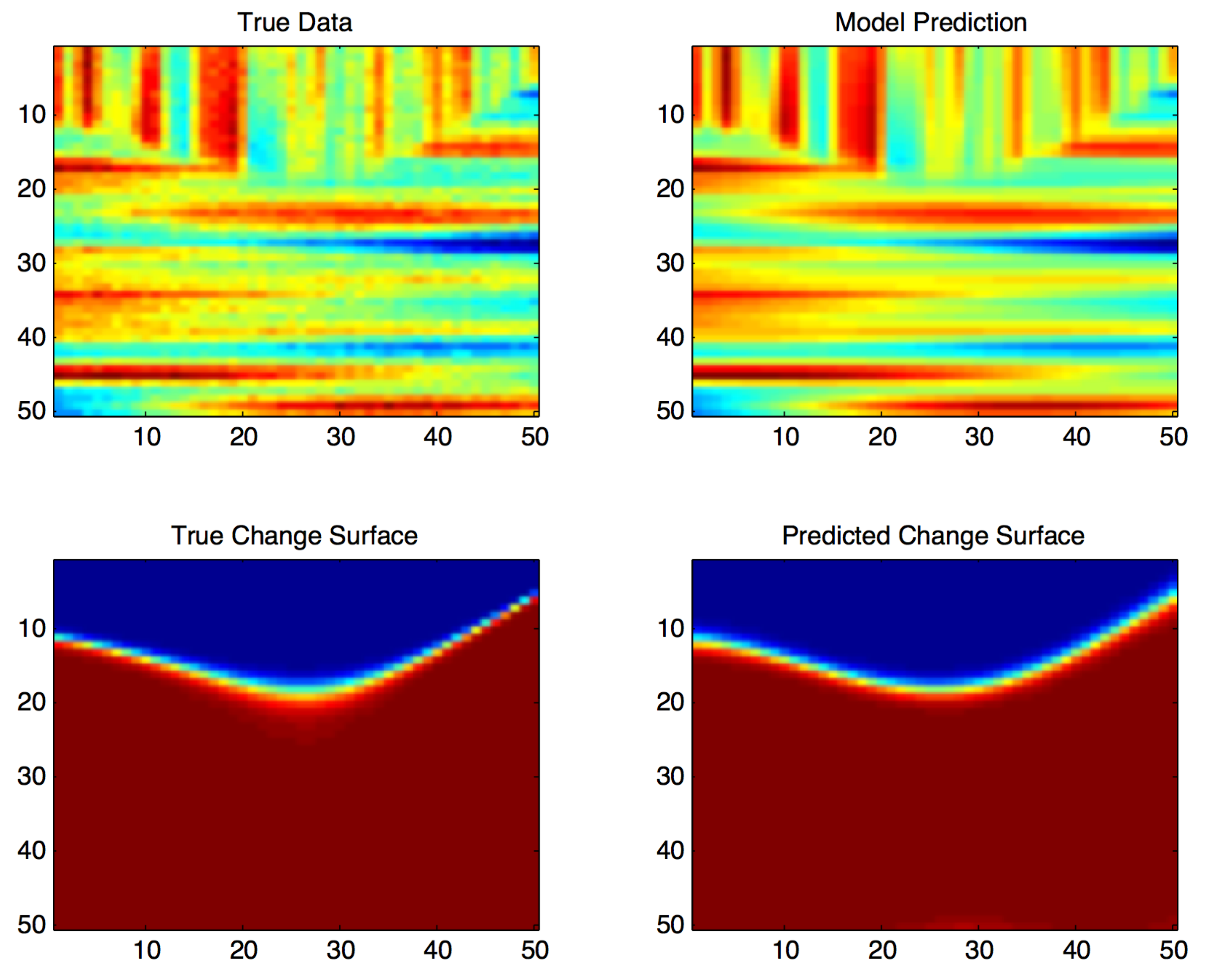}
\caption{Numerical data experiment. The top-left depicts the data; the bottom-left shows the true change surface with the range from blue to red depicting $\sigma(w_1(x))$. The top-right depicts the predicted output; the bottom-right shows the predicted change surface.}
\label{fig:synthetic1122222295}
\end{centering}
\end{figure}

\begin{figure}[h]
\begin{centering}
 \includegraphics[width=0.6\textwidth]{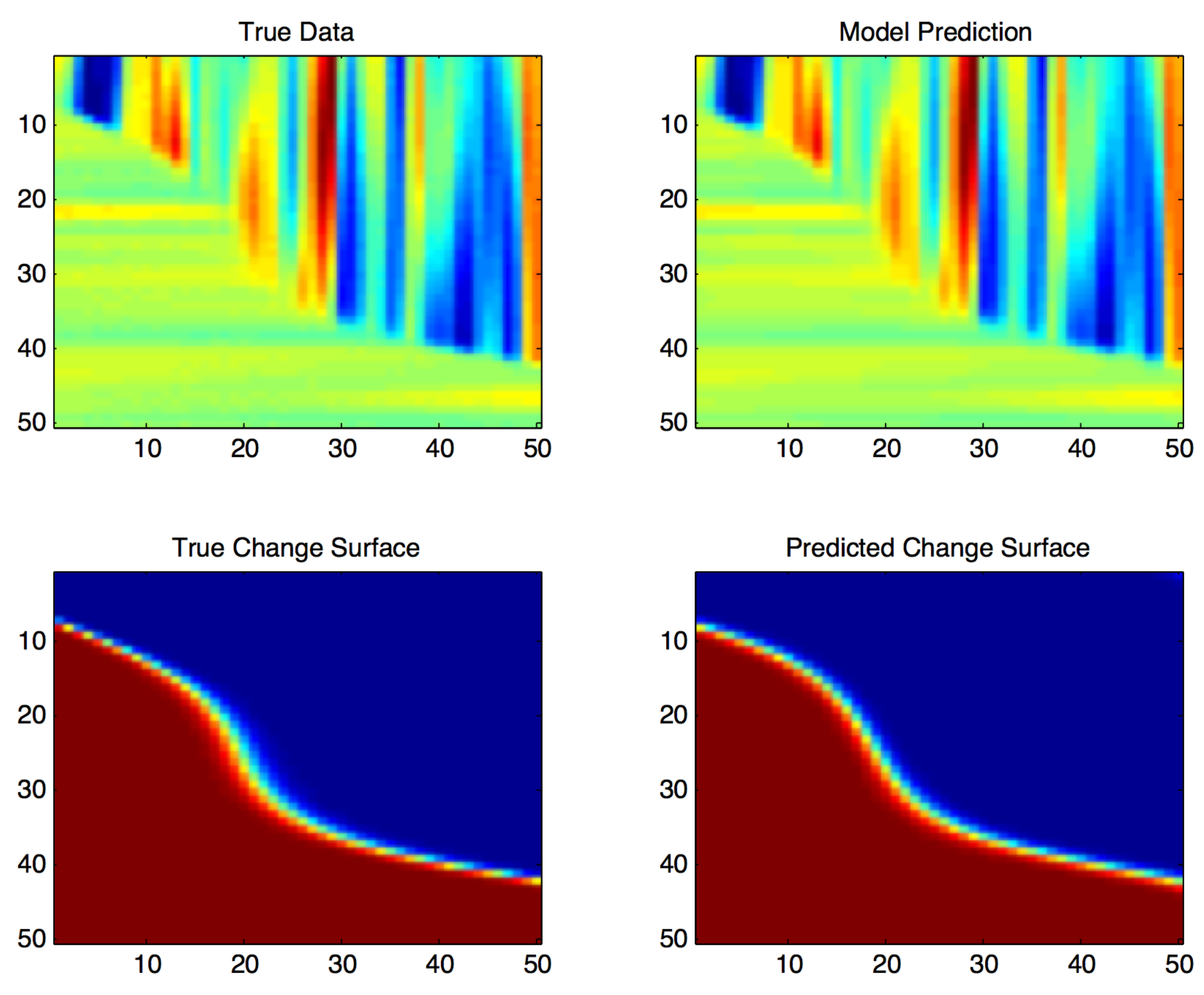}
\caption{Numerical data experiment. The top-left depicts the data; the bottom-left shows the true change surface with the range from blue to red depicting $\sigma(w_1(x))$. The top-right depicts the predicted output; the bottom-right shows the predicted change surface.}
\label{fig:synthetic11222222}
\end{centering}
\end{figure}

We apply our change surface model with two latent functions, spectral mixture kernels, and $w(x)$ defined by 5 RKS features. We do not provide the model prior information about the change surface or latent functions. Figures \ref{fig:synthetic1122222295} and \ref{fig:synthetic11222222} depict typical results using the initialization procedure followed by analytic optimization. The model captures the change surface and produces an appropriate regression over the data.

Using synthetic data, we create a predictive test by splitting the data into training and testing sets. We compare our smooth change surface model to three other expressive, scalable methods: sparse spectrum Gaussian process with 500 basis functions \citep{lazaro2010sparse}, sparse spectrum Gaussian process with fixed spectral points with 500 basis functions \citep{lazaro2010sparse}, and a Gaussian process with multiplicative spectral mixture kernels in each dimension. For each method we average the results for 10 random restarts. Table \ref{tab:comparison} shows the normalized mean squared error (NMSE) of each method, where $\bar{y}_{train}$ is the mean of the training data. 
\begin{eqnarray}
\label{eq:NMSE}
\text{NMSE} = \frac{\| y_{test} - y_{pred} \|_2^2}{\| y_{test} - \bar{y}_{train} \|_2^2}
\end{eqnarray}

\begin{table}[]
\centering
\caption{Comparison of prediction using flexible, scalable Gaussian process methods on synthetic multidimensional change-surface data. }
\vspace{0.3in}
\label{tab:comparison}
\begin{tabular}{|l|l|}
\hline
\textbf{Method}       & \textbf{NMSE} \\ \hline \hline
Smooth change surface    & 0.00078       \\ \hline
SSGP       & 0.01530       \\ \hline
SSGP fixed  & 0.02820       \\ \hline
Spectral mixture      & 0.00200       \\ \hline
\end{tabular}
\end{table}

Our change surface model performed best due to the expressive nonstationary covariance function that fits to the different functional regimes in the data. Although the alternate methods can flexibly adapt to the data, they must account for the change in covariance structure by setting an effectively shorter length-scale over the data. Thus their predictive accuracy is reduced compared to the change surface model.

\subsection{British Coal Mining Data}
\label{coal_exp}

British coal mining accidents from 1861 to 1962 have been well studied in the point process and changepoint literature \citep{raftery1986bayesian, adams2007bayesian}. We use yearly counts of accidents from \citet{carlin1992hierarchical}. Domain knowledge suggests that the Coal Mines Regulation Act of 1887 affected the underlying process of coal mine accidents. This act limited child labor in mines, detailed inspection procedures, and regulated construction standards.

\begin{figure}[h]
\begin{centering}
 \includegraphics[width=0.7\textwidth]{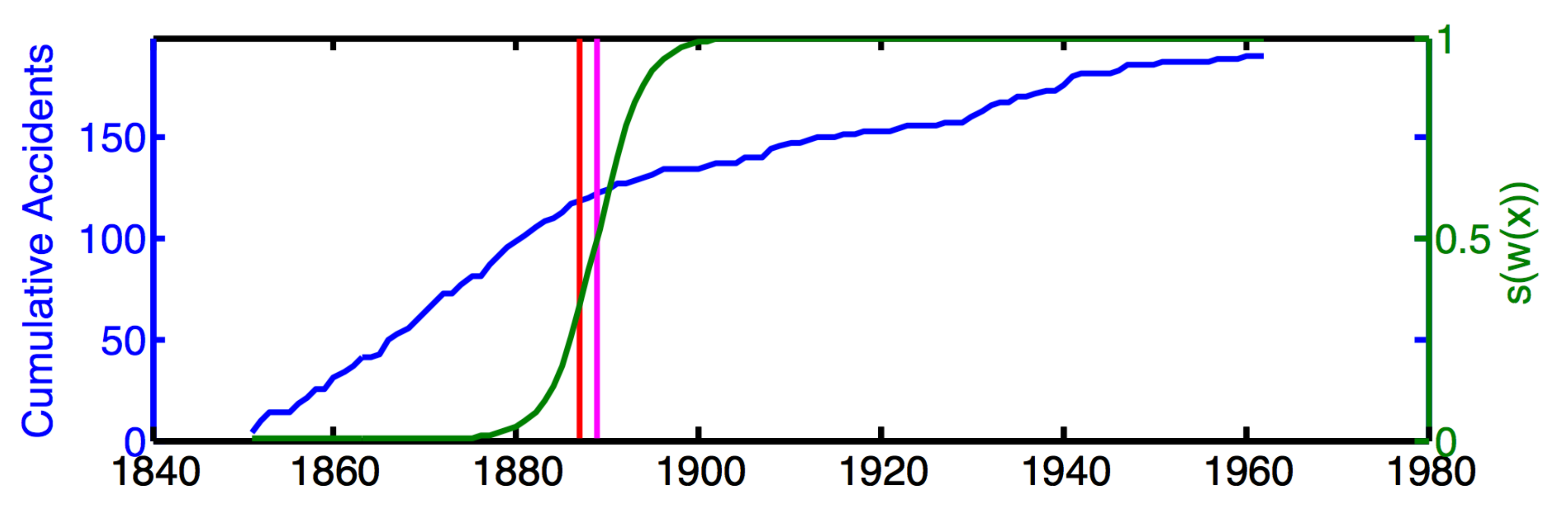}
\caption{British coal mining accidents from 1851 to 1962. The blue line depicts cumulative annual accidents, the green line plots $\sigma(w(x))$, the vertical red line marks the Coal Mines Regulation Act of 1887, and the vertical magenta line indicates $\sigma(w_1(x))=0.5$.}
\label{fig:coal_results}
\end{centering}
\end{figure}

We apply our change surface model with two latent functions, spectral mixture kernels, and $w(x)$ defined by 5 RKS features. We do not provide the model with prior information about the 1887 legislation date. Figure \ref{fig:coal_results} depict the cumulative data and predicted change surface. The red line marks the year 1887 and the magenta line marks $x:\sigma(w(x))=0.5$. Our algorithm correctly identified the change region and suggests a gradual change that took 11.3 years to transition from $\sigma(w_1(x))=0.1$ to $\sigma(w_1(x))=0.9$.

\subsection{US Disease Data}
\label{sec:disease_exp}
Measles was nearly eradicated in the United States following the introduction of the measles vaccine in 1963. We analyze monthly incidence data for measles from 1935 to 2003 in each of the continental United States and the District of Columbia, made publicly available by Project Tycho \citep{van2013contagious}. We fit the model to $\approx 33,000$ data points where $x\in R^{3}$ with two spatial dimensions representing centroids of each state and one temporal dimension.

We apply our change surface model with two latent functions, spectral mixture kernels, and $w(x)$ defined by 5 RKS features. We do not provide prior information about the 1963 vaccination date.

\begin{figure}[h]
\begin{centering}
 \includegraphics[width=0.7\textwidth]{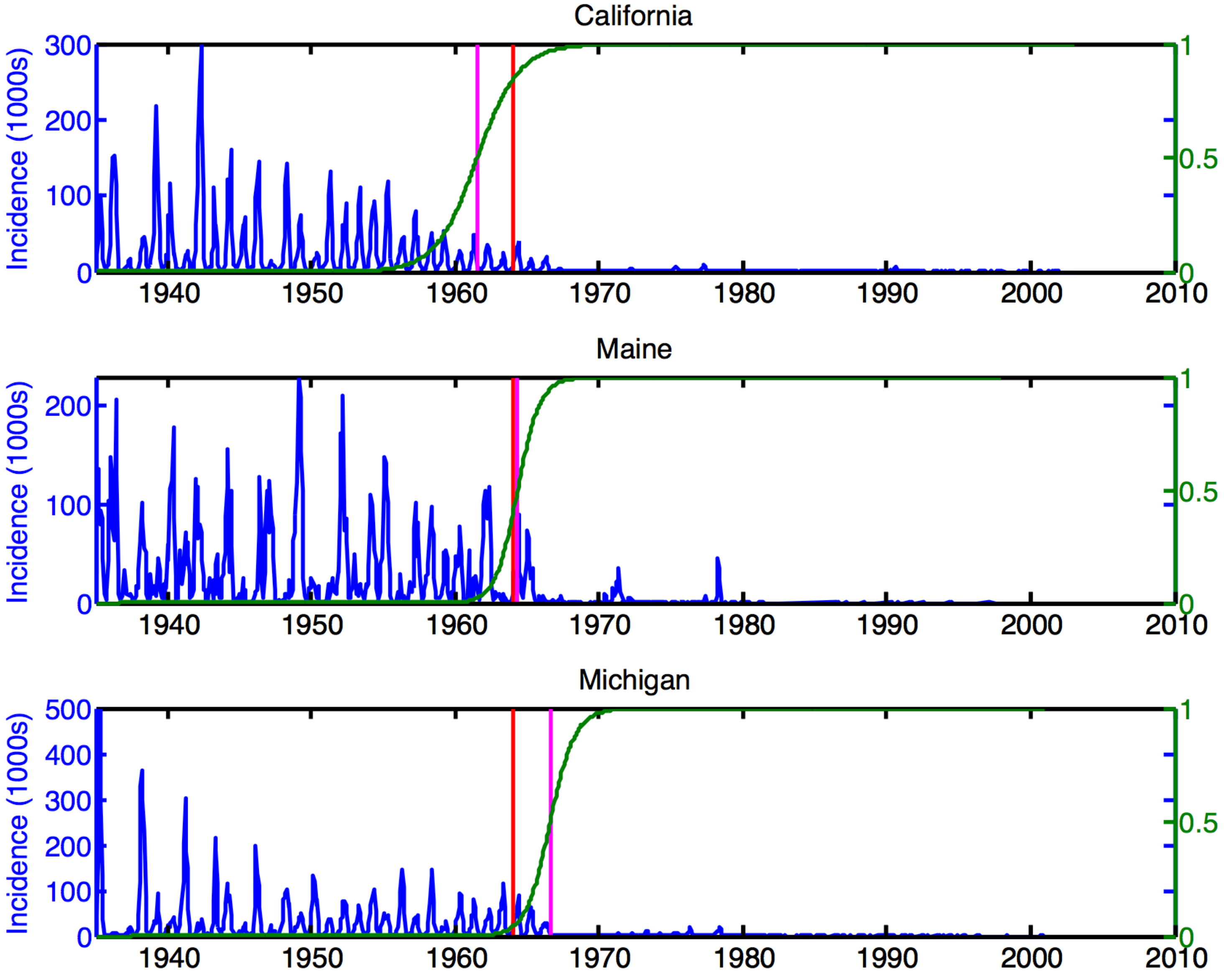}
\caption{Measles incidence levels from 3 states, 1935 - 2003. The green line plots $\sigma(w(x_{state}))$, the vertical red line indicates the vaccine in 1963, and the magenta line indicates $\sigma(w(x_{state}))=0.5$.}
\label{fig:tycho_states3}
\end{centering}
\end{figure}

Results for three states are shown in Figure \ref{fig:tycho_states3} along with the predicted change surface. The red line marks the vaccine year of 1963, while the magenta line marks the points where $\sigma(w(x_{state}))=0.5$. Our algorithm correctly identified the time frame when the measles vaccine was released in the US.

\begin{figure}[h]
\begin{centering}
 \includegraphics[width=0.7\textwidth]{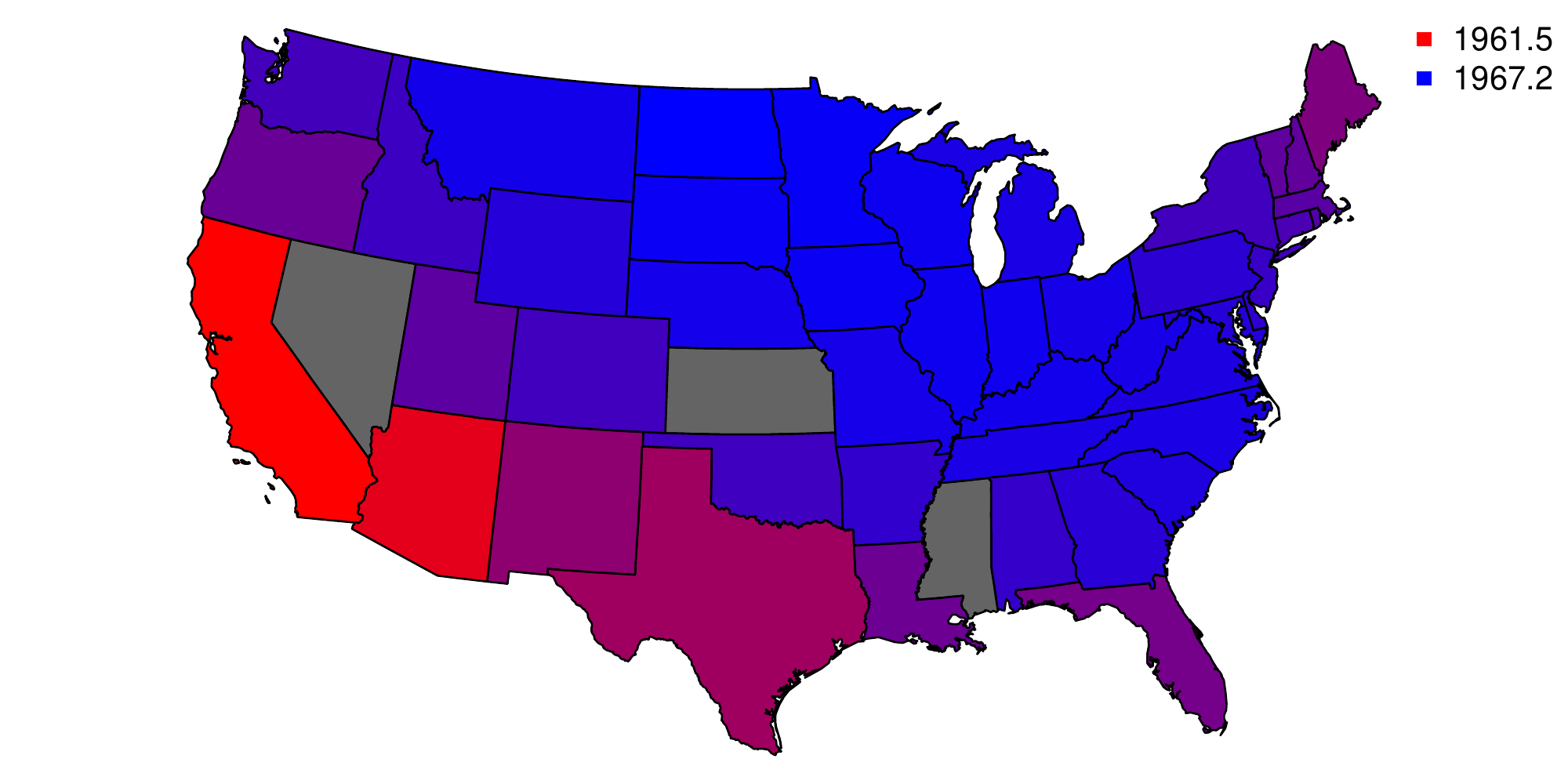}
\caption{US states colored by the date where $\sigma(w(x_{state}))=0.5$. Red indicates earlier dates, with California being the earliest. Blue indicates later dates, with North Dakota being the latest. Grayed out states were missing in the dataset.} 
\label{fig:US_dates}
\end{centering}
\end{figure}
\begin{figure}[h]
\begin{centering}
 \includegraphics[width=0.7\textwidth]{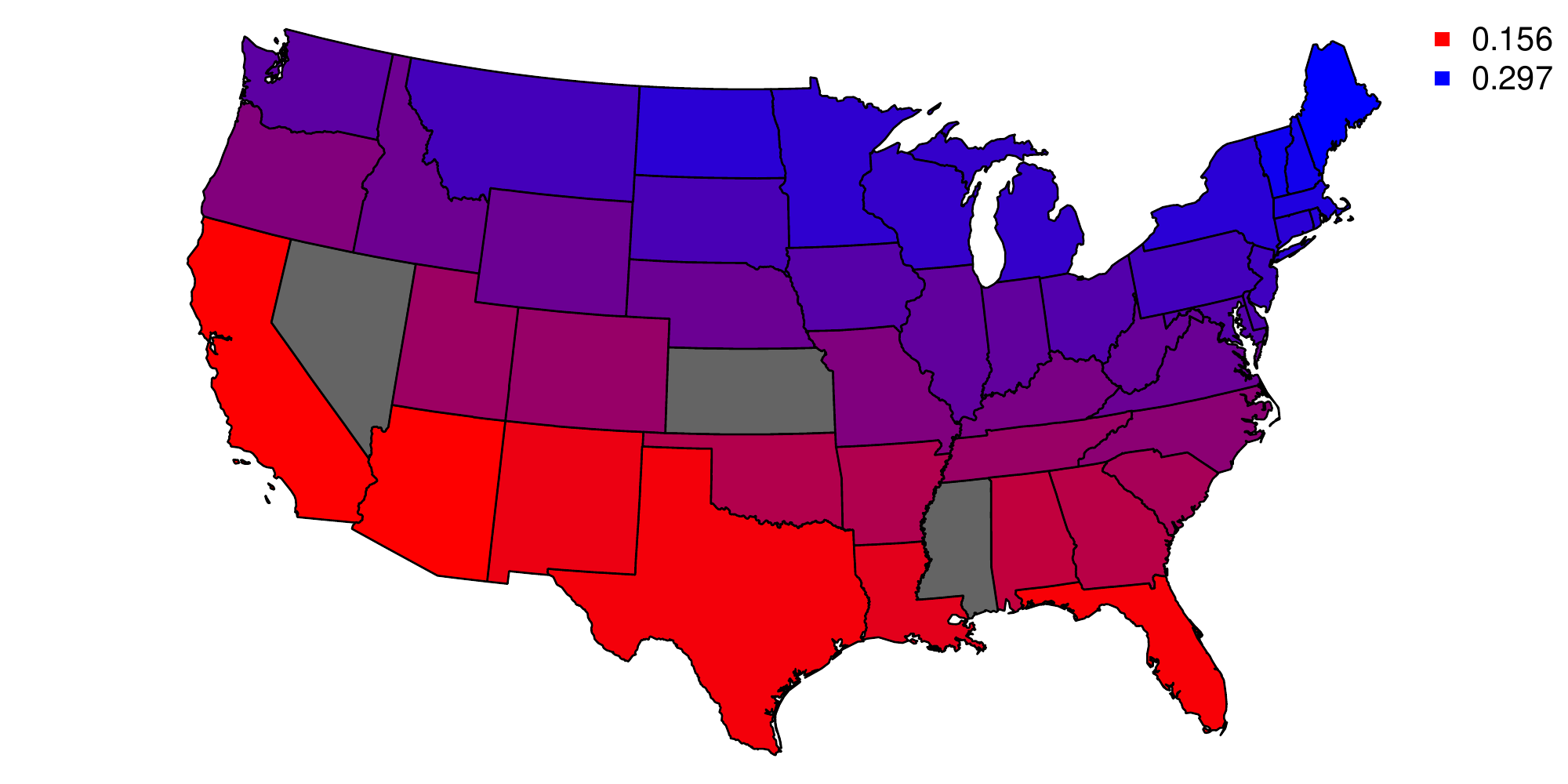}
\caption{US states colored by the slope of $\sigma(w(x_{state}))$ from $0.25$ to $0.75$. Red indicates flatter slopes, with Arizona being the lowest. Blue indicates steeper slopes, with Maine being the highest. Grayed out states were missing in the dataset.}
\label{fig:US_slope}
\end{centering}
\end{figure}

Additionally, the model suggests that the effect of the measles vaccine varied both temporally and spatially. In Figure \ref{fig:US_dates} we depict the midpoint, $\sigma(w(x_{state}))=0.5$, for each state. We discover that there is an approximately 6 year difference in midpoint between states. 
In Figure \ref{fig:US_slope} we depict the change surface slope from $\sigma(w(x_{state}))=0.25$ to $\sigma(w(x_{state}))=0.75$ for each state to estimate the rate of change. Here we find that some states had approximately twice the rate of change as others.
These variations in the change surface illustrate how the measles vaccine affected states heterogeneously over space and time. They suggest that further scientific research is warranted to understand the underlying causes of this heterogeneity in order to provide insight for future vaccination programs.

\section{Conclusions}
\label{sec:conclusions}

We presented a scalable, multidimensional Gaussian process model with expressive kernel structure which can learn a complex change surface from data. Using the Weyl inequality, we perform efficient inference with additive kernel structure using Kronecker methods, enabling a multidimensional non-separable kernel. Additionally, we introduce a novel initialization algorithm for learning the $w(x)$ RKS features and spectral mixture kernels. Finally, we apply our model to numerical and real world data, illustrating how it can characterize heterogeneous spatio-temporal change surfaces, yielding scientifically relevant insights.

The work on changepoint modeling is extensive and the current work cannot address all facets of the literature. Future work can extend our retrospective analysis to address sequential change surface detection. Additionally, the current method can be extended to automatically determining the number of latent functions using a automatic modeling discovery approach such as \citet{lloyd2014automatic}.

\subsubsection*{Acknowledgements}
This material is based upon work supported by the National Science Foundation Graduate Research Fellowship under Grant No. DGE 1252522 and the National Science Foundation award No. IIS-0953330.

\clearpage

\bibliographystyle{apalike}
\bibliography{GPCS_refs}

\end{document}